# Intelligent Pathological Diagnosis of Gestational Trophoblastic Diseases via Visual-Language Deep Learning Model

## Author Information

Yuhang Liu[1,a], Yueyang Cang[1,a], Wenge Que [1,f], Xinru Bai[1,b], Xingtong Wang[a], Kuisheng Chen[c], Jingya Li[d], Xiaoteng Zhang[a], Xinmin Li[e], Lixia Zhang[f], Pingge Hu[g], Qiaoting Xie[h], Peiyu Xu[i], Xianxu Zeng[b,*], Li Shi[a,*]

[a]*Department of Automation, Tsinghua University, Beijing, China*

[b]*Department of Pathology, Third Affiliated Hospital of Zhengzhou University, Zhengzhou, China*

[c]*Department of Pathology, The First Affiliated Hospital of Zhengzhou University, Zhengzhou, China*

[d]*Luoyang Maternal and Child Health Care Hospital, Luoyang, China*

[e]*Department of Pathology, Women and Infants Hospital of Zhengzhou, Zhengzhou, China*

[f]*Xinmi Maternal and Child Health Care Hospital, Xinmi, China*

[g]*China Academy of Information and Communications Technology, Beijing, China*

[h]*Anyang Maternal and Child Health Care Hospital, Anyang, China*

[i]*Suixian Maternal and Child Health Care Hospital, Shangqiu, China*

[f]*College of Materials Science and Engineering, Donghua University, Shanghai, China*

\* *Corresponding authors.*

[1] *These authors contributed equally to this work.*

# Abstract

The pathological diagnosis of gestational trophoblastic disease(GTD) takes a long time, relies heavily on the experience of pathologists, and the consistency of initial diagnosis is low, which seriously threatens maternal health and reproductive outcomes. We developed an expert model for GTD pathological diagnosis, named GTDoctor. GTDoctor can perform pixel-based lesion segmentation on pathological slides, and output diagnostic conclusions and personalized pathological analysis results. We developed a software system, GTDiagnosis, based on this technology and conducted clinical trials. The retrospective results demonstrated that GTDiagnosis achieved a mean precision of over 0.91 for lesion detection in pathological slides (n=679 slides). In prospective studies, pathologists using GTDiagnosis attained a Positive Predictive Value of 95.59% (n=68 patients). The tool reduced average diagnostic time from 56 to 16 seconds per case (n=285 patients). GTDoctor and GTDiagnosis offer a novel solution for GTD pathological diagnosis, enhancing diagnostic performance and efficiency while maintaining clinical interpretability.

# Introduction

Gestational trophoblastic diseases (GTDs) are a group of disorders originating from placental trophoblastic cells. According to the World Health Organization (WHO) 2020 (5th edition) classification of female genital system tumors[1], GTDs can be histologically divided into: (1) hydatidiform moles (HM), (2) Gestational trophoblastic neoplasia (GTN), including choriocarcinoma, etc, (3) tumor-like lesions, (4) abnormal (non-molar) villous lesions. GTDs pose a significant potential threat to maternal health. If common HMs are not properly identified and managed, they may progress into life-threatening choriocarcinoma. Therefore, early diagnosis of GTDs is of critical clinical significance[2,3].

Pathological examination is the gold standard for GTD diagnosis. Currently, there are three main clinical pathological diagnostic methods: serum human chorionic gonadotropin (hCG) testing, short tandem repeat (STR) testing, and artificial diagnosis through microscopic examination of tissue sections[4-7]. Among these, microscopic histopathological examination is cost-effective and widely used. However, there are several challenges in pathological diagnosis. HCG testing typically requires multiple dynamic monitoring sessions over an extended period, leading to delays for patients who need rapid diagnosis and timely intervention. STR testing, on the other hand, has a high technical threshold, and non-specialized hospitals may lack the necessary facilities[6,25-27]. Moreover, manual diagnosis is inherently subjective and heavily reliant on the expertise of experienced pathologists, who are currently in short supply due to a significant shortage of qualified professionals. In China, there are fewer than 20,000 registered pathologists[8], with 70% concentrated in top-tier hospitals. For many obstetrics and gynecology hospitals, there is no dedicated pathology department, making pathological examination inconvenient, as tissue samples need to be sent to other larger hospitals for testing. This underscores the urgent need for reliable intelligent diagnostic support tools to improve the efficiency and accuracy of pathological diagnoses in this field[28-30].

Deep learning and large-model technologies have introduced promising solutions for pathological diagnosis [19-24]. Generic vision-language models for pathology are able to analyze the characteristics of various types of slides[9-11,37], while multimodal large language models have the capabilities of providing analytical insights from image and text data[12-14], and heuristic neural networks effectively extract features of hydatidiform moles[15,16]. These advances are encouraging but still facing notable limitations. Currently, most models tend to focus more on the comprehensive analysis of pathology slides for general diseases, while diseases related to gynecology, such as GTD, are often overlooked. There are no large models currently using GTD-related data for learning and training. Furthermore, artificial intelligence and large model technologies are still far from being fully applied in practice, with very few models addressing real clinical situations or considering the clinical needs under different medical conditions. To address these challenges, we have collected the first GTD database and developed a multi-scenario GTD diagnostic expert model, GTDoctor, which has undergone complete clinical application. Our research successfully bridges the gap for large-model deployment and clinical application.

In this study, we first established a multi-center pathological slide database for GTDs. Based on patients' pathological images, diagnostic guidelines, and expert pathologists' knowledge, we developed developed an expert model for GTD pathological diagnosis, GTDoctor, and its corresponding intelligent diagnostic support software system, GTDiagnosis, which is applicable to multiple scenarios. This model can self-learn and adapt to different slide styles based on regional data variations, making it suitable for use with both microscopes and slide scanners in hospitals of varying medical levels. We conducted retrospective and prospective trials across seven centers, including the Third Affiliated Hospital of Zhengzhou University, to assess improvements in diagnostic accuracy and efficiency. The results were remarkable, demonstrating that our system can greatly assist pathologists in clinical diagnosis and research within the GTD field. Fig. 1 presents the system architecture diagram of our project.

## Results

### Data and patient overview

This study was approved by the Institutional Review Boards of the Third Affiliated Hospital of Zhengzhou University, the First Affiliated Hospital of Zhengzhou University, the Luo Yang Maternal and Child Health Care Hospital, the Xinmi Maternal and Child Health Care Hospital, the Anyang Maternal and Child Health Care Hospital, the Sui County Maternal and Child Health Care Hospital, and the Women and Infants Hospital of Zhengzhou. In this study, we used a total of 1,916 Whole Slide Images (WSI) from 831 patients across the seven centers, along with corresponding baseline information and clinical cases for each patient. All patient samples were collected between 2015 and 2024.

The data from these seven centers were organized into four major cohorts and nine sub-cohorts. Cohort A, from the Third Affiliated Hospital of Zhengzhou University, was divided into Cohort A-1 (406 patients, 1038 slides) and Cohort A-2 (112 patients, 253 slides). Cohort A-1 was used for model training and testing, while Cohort A-2 was used for internal retrospective validation. Cohort B consisted of data from five medical centers and was used for external retrospective validation to evaluate the generalization performance of GTDiagnosis. These included Anyang Maternal and Child Health Care Hospital (Cohort B-1, 22 patients, 118 slides), Xinmi Maternal and Child Health Care Hospital (Cohort B-2, 21 patients, 96 slides), the Women and Infants Hospital of Zhengzhou (Cohort B-3, 80 patients, 80 slides), the First Affiliated Hospital of Zhengzhou University (Cohort B-4, 63 patients, 111 slides), and Sui County Maternal and Child Health Care Hospital (Cohort B-5, 19 patients, 21 slides). Cohort C, from Luoyang Maternal and Child Health Care Hospital, was used to quantify the impact of GTDiagnosis on traditional diagnostic practices (Cohort C-1, 40 patients, 72 slides). Additionally, a prospective testing cohort from a multi-center mixed data source was utilized (Cohort D-1, 68 patients, 127 slides). There was no overlap of patients between any training and testing sets for different tasks. Table 1 shows the baseline characteristics of the data.

## Data Cleaning and Dataset Construction

**Data Selection:** We collaborated with pathologists to first filter the WSI data, removing slides with issues such as scanning blur, air bubbles, tissue folding, and abnormal staining to ensure that the data used for training were of high quality. We obtained digital pathology slide images captured under a 40x objective lens, with a resolution typically around 60,000 × 60,000 pixels.

**Data Annotation:** We performed annotation on the filtered pathology slides. Pathologists used a Windows-based annotation system to label edema lesions, hyperplasia lesions, and villi regions at the pixel level, using different colors for each category. The pathology slides to be annotated were randomly assigned to a pathologist for labeling. Once the annotation was completed, the labeled slide was transferred to another pathologist for verification. In cases where consensus could not be reached, a third senior pathologist with over 15 years of experience would make the final judgment. In summary, the annotation process consisted of initial labeling, verification, and potential review stages.After accumulating a certain number of annotated slides (237 WSI), we improved the annotation workflow to enhance efficiency. We used these annotated data to train a U-Net-based semantic segmentation network for basic lesion segmentation[17,18]. This model was then used to segment unlabeled slides, which helped reduce the need for initial manual annotation. The majority of lesions could be correctly annotated by the model, greatly alleviating the workload and time required from pathologists. The U-Net model's annotated results were randomly assigned to a pathologist for revision, then passed to another pathologist for validation. In cases of disagreement, a third senior pathologist with over 15 years of experience would make the final judgment. The process thus consisted of initial model annotation, revision and validation, and potential review.

**Dataset Construction:** In constructing the pathology slide dataset, we divided each WSI into a series of 512 × 512 patches at a 4× magnification level. The patches were adjacent to each other, covering the entire WSI. We then used an adaptive variance threshold algorithm to filter out patches dominated by slide background, retaining only images with valid tissue. In total, we extracted 276,562 patches from the 1,916 WSI.

## Multicenter Validation of GTDiagnosis

We evaluated the performance of GTDiagnosis across six medical centers (Cohorts A-2, Cohorts B-1 to B-5). For these patients, pathological slides were digitized using whole-slide scanners and served as the input for the model. We assessed GTDiagnosis on two tasks: lesion segmentation and disease diagnosis. The ROC curves for edema and hyperplasia lesions are shown in Fig. 3. For edema lesion segmentation, the AUC ranged from 0.968 to 0.998 across different medical centers, while for hyperplasia lesions, the AUC ranged from 0.906 to 0.969. These results demonstrate that our model achieved consistent and high segmentation performance across data from multiple hospitals.

Detailed performance metrics for each cohort are presented in Fig. 3. We evaluated mIoU, mDice, mRecall, and mPrecision. For the cohort from the Third Affiliated Hospital of Zhengzhou University (Cohort A-2), the model achieved mPrecision = 0.969 (95% CI: 0.961–0.977), mRecall = 0.947 (95% CI:

0.940–0.954), mDice = 0.966 (95% CI: 0.958–0.974), and mIoU = 0.955 (95% CI: 0.946–0.964). Across all centers, mPrecision, mRecall, and mDice were consistently above 0.9, while mIoU was above 0.84. These results highlight the strong accuracy and excellent generalizability of our vision model. During training, we prioritized maintaining a high Recall, as this reflects the model's ability to detect potential lesions—a critical capability for assisting clinicians in diagnosis and decision-making.

The diagnostic results for each cohort are summarized in confusion matrices, as shown in Fig. 3. For Cohort A-2, the accuracy was 0.91; at other medical centers, our model also achieved significant results, with Cohort B-1 at 0.95, Cohort B-2 at 0.86, Cohort B-3 at 0.86, Cohort B-4 at 0.84, and Cohort B-5 at 0.74. Overall, the mean recall for hydatidiform mole was 0.90, while the mean recall for choriocarcinoma was 0.86. These findings indicate that our model is highly sensitive in identifying both diseases. Additionally, it is worth noting that these results were obtained using GTDiagnosis alone. In actual clinical use, GTDiagnosis serves as an auxiliary tool for doctors, with the final decision on contentious pathology resting with the clinicians. Therefore, the accuracy in real-world applications is expected to be higher than these values.

To evaluate the lesion segmentation performance of our model in microscopic fields, we selected 20 pathological slides from hydatidiform mole patients in the cohort from the Third Affiliated Hospital of Zhengzhou University (Cohort A-2). The slides were examined under a 4× magnification microscope, with the slides manually moved by a pathologist. During this process, GTDiagnosis performed real-time lesion recognition and segmentation on the microscopic field images and saved the results for subsequent stitching. Once the entire slide was scanned under the microscope, a stitched image containing the model's segmentation results for edema and hyperplasia lesions was generated. These stitched images were then registered and compared with the ground truth annotations provided by the pathologists to validate the model's performance in microscopic fields.

The lesion segmentation performance of GTDiagnosis in microscopic fields achieved mPrecision = 0.825, mRecall = 0.842, mDice = 0.838, and mIoU = 0.775. While the performance was slightly lower than that on digitized pathology slides, likely due to factors such as lighting variations and slide quality, the model still demonstrated the ability to detect most lesions effectively. This indicates that GTDiagnosis can serve as a valuable auxiliary tool in assisting pathologists during microscopic diagnosis, particularly for less experienced pathologists.

## Improvement in Clinical Efficiency and Diagnostic Accuracy

To assess the practical impact of our model on pathologists' daily diagnoses, we conducted a validation experiment using data from the Luoyang Maternal and Child Health Hospital (Cohort C-1). Two pathologists were recruited for the test, categorized as junior pathologists (0-3 years of experience) and senior pathologists (3-6 years of experience). For each pathologist, 20 cases and their sections were randomly selected for analysis and diagnosis. The results demonstrated significant advantages in both diagnostic time and accuracy when using GTDiagnosis compared to the traditional microscope-based diagnostic process.

For junior pathologists, the average diagnosis time per patient decreased from 57.35 seconds to 14.58 seconds, and the average diagnostic accuracy improved from 82.88% to 91.10%. For senior pathologists, the average diagnosis time per patient decreased from 54.66 seconds to 17.83 seconds, with diagnostic accuracy increasing from 95.21% to 98.63%. In the test, the average diagnosis time refers to the time taken by the doctor from receiving a patient's pathological slide to completing the diagnosis, while the average diagnostic accuracy is the proportion of correct diagnostic conclusions made by the doctor after analyzing the pathological slide. The experimental group was analyzed by the microscope equipped with GTDiagnosis, while the control group was analyzed by the traditional microscope.

These results indicate that GTDiagnosis can reduce the average diagnostic time to about a quarter of the original time, while simultaneously improving diagnostic accuracy. This has the potential to significantly alleviate pathologists from repetitive tasks, particularly for junior pathologists, as the segmentation results from the visual model provide strong diagnostic support. In addition, by using the combination of pathological section scanner and GTDiagnosis system, doctors can directly analyze the visual field of the whole section, which will further improve efficiency. This proves the remarkable time efficiency advantage of our method.

## Prospective Clinical Testing of GTDiagnosis

In the prospective testing phase, GTDiagnosis was utilized to assist pathologists in completing diagnostic tasks, and its performance was meticulously documented. This study was conducted from May 2024 to October 2024 at multiple centers. A total of 68 patients and 127 pathological sections were included. Of these, 21 patients and 45 slides were from the Third Affiliated Hospital of Zhengzhou University; 19 patients and 34 slides were from Luoyang Maternal and Child Health Hospital; 9 patients and 12 slides were from Xinmi Maternal and Child Health Hospital; 11 patients and 23 slides were from the Women and Infants Hospital of Zhengzhou; and 8 patients and 13 slides were from Anyang Maternal and Child Health Hospital.

In the process, tissue sections were first batch-scanned into digital slides using a pathological section scanner, and then pathologists conducted online diagnoses on computer devices equipped with GTDiagnosis. Finally, GTDiagnosis integrated and archived the diagnostic results, achieving a full-process intelligent auxiliary diagnosis of GTD. For cases diagnosed as abnormal, further pathological tests were conducted, including hCG testing and STR testing. We considered these test results as ground truth. Based on the further testing results, we calculated the Positive Predictive Value, which refers to the initial diagnosis accuracy. The results indicated that the Positive Predictive Value of pathologists using GTDiagnosis reached 95.59% (65 out of 68), demonstrating that pathologists could achieve highly accurate diagnoses based solely on image data extracted from pathological sections using GTDiagnosis.

## **Discussion**

In this study, we present a multimodal AI-based diagnostic support system for Gestational Trophoblastic Diseases (GTDs), called GTDiagnosis. This system automates the analysis of patient endometrial pathological slides and generates diagnostic results and reports. GTDiagnosis leverages a multicenter dataset consisting of 1,916 whole-slide pathological images (276,562 patches), integrated with state-of-the-art visual and language models. Compared to traditional manual diagnostic methods, GTDiagnosis shows significant improvements in both diagnostic accuracy and efficiency for GTDs.

In the field of artificial intelligence-assisted GT diagnosis, such as the work by Patison Palee, used 939 hydatidiform mole pathology slides to extract 15 numerical features for the identification of proliferative tissue, and the model was limited to classifying the types of hydatidiform moles without clinical testing[15]. Moreover, there is no large-scale, multicenter study focusing on lesion detection and intelligent diagnostic support for GTDs. Our work provides a more comprehensive and detailed solution for the computer aided diagnosis (CAD) of GTDs. The novelties of our proposed GTDiagnosis can be summarized as follows: (1) Comprehensive Diagnostic Capability: GTDiagnosis combines lesion segmentation, classification, and pathology analysis to provide end-to-end diagnostic support for multiple GTD types. In external validation datasets from six independent medical centers, the model demonstrated excellent performance in lesion segmentation, with mPrecision, mRecall, and mDice exceeding 0.90, showcasing its remarkable generalization ability. In the disease classification task, the average recall for hydatidiform mole was 0.90, and for choriocarcinoma, it was 0.86, indicating high sensitivity to key GTD types. In addition, the system has the ability to perform real-time analysis of the microscopic field of view, enabling its application in resource-limited settings without digital pathology scanning equipment. (2) Integration of Multimodal Features: GTDiagnosis effectively integrates text and image information to perform lesion segmentation and disease classification. The multi-scale adaptive attention vision model developed in this study significantly enhances segmentation accuracy and improves recognition of multi-scale lesions in GTDs. Additionally, the decision model incorporates pathologists' expertise and clinical guidelines, enabling the system to provide authoritative and professional classification results. Compared to single-modality image segmentation, we have introduced rich textual knowledge to analyze patients' pathological slide images. Compared to black-box pure neural network methods, our approach enhances the clinical interpretability of AI-based diagnosis. (3) Diagnostic Support and Efficiency Improvement: GTDiagnosis introduces advanced language modeling techniques to generate in-depth diagnostic reports.

By using retrieval-augmented generation (RAG), the system can retrieve relevant medical literature and generate personalized, evidence-based diagnostic insights. (4) User-Friendliness and Self-Evolution Potential: Pathologists can master the system within 20 minutes, with significant improvements in workflow efficiency. The results show that GTDiagnosis effectively reduces pathologists' diagnostic workload and enhances diagnostic confidence. Additionally, the system's online learning capability allows newly collected data to be used for model retraining, which is beneficial for its deployment across multiple centers and indicates its strong evolution potential.

It is noteworthy that AI-assisted diagnosis faces many challenges in clinical practice. When using visual models to analyze pathology slide images, challenges arise due to variations in staining methods, slide scanning instruments, and regional differences between hospitals. These factors can complicate the analysis performed by the visual models. To address the issue of inconsistent slide styles and scales, we have made several efforts. During the model training phase, we applied data augmentation techniques such as color enhancement and size scaling. Additionally, we targeted improvements to traditional attention mechanisms, allowing the modified algorithm to perform adaptive attention matching across multiple scales—from micro to macro—enabling it to better accommodate a wider variety of slide styles. Finally, we incorporated an online learning algorithm, enabling the model to evolve autonomously. After deploying our base model in different regions, local slide data can be used to fine-tune the model, further improving its adaptability. These efforts have yielded positive results in multi-center retrospective studies, and prospective experiments have also demonstrated the model's reliability in clinical applications.

Moreover, we have thoroughly considered the practical demands of medical applications. In low-tolerance medical scenarios, we combine structured decision models with unstructured large language models for pathology analysis. This hybrid approach mitigates the hallucination effects of large models by incorporating a knowledge framework that constrains them. The structured decision model integrates the diagnostic logic and experience of medical professionals, while the large language model draws on an extensive knowledge base to provide expert generative analysis. Together, they form a robust system that ensures both reliability and precision in complex medical environments.

Although GTDiagnosis shows promising results, there are some limitations. First, while the system performs excellently on digital pathology slides, its performance in microscopic fields is slightly affected by factors such as lighting and slide quality. Second, the online learning capability of the system requires strict monitoring to prevent overfitting and ensure model stability during continuous updates. In the future, we hope to use more advanced multimodal large models to further improve GTDiagnosis's predictive performance, especially in identifying complex samples. Additionally, we can leverage the scalability of our workflow to include more related diseases, thereby enhancing the model's applicability. In conclusion, our research has reduced the lengthy manual diagnostic process to mere seconds, while also effectively assisting doctors in analysis and decision-making, particularly providing a powerful tool for junior pathologists with limited experience. In regions with limited medical resources, our system can be accessed via a microscope for auxiliary analysis, while hospitals equipped with the appropriate conditions can utilize pathology slide scanners for batch analysis, thereby digitizing and automating the manual examination process. As an intelligent diagnostic support system, GTDiagnosis can offer valuable assistance to pathologists and researchers in clinical practice, medical education, and research, providing robust support for intelligent and personalized diagnoses.

## Methods

## Lesion Segmentation Visual Model

The lesion segmentation process consists of two main parts: first, the segmentation of the villous areas, and second, the segmentation of the lesion areas. The model is built based on a two-stage training process. In the first stage, the network first segments the villous areas, outputting a probability map for each pixel indicating its likelihood of belonging to the corresponding villous region. Since lesions are typically located within the villous areas, these probability maps provide crucial information for lesion segmentation. To effectively utilize this information, the input for lesion segmentation includes not only the original pathological RGB images but also the villous segmentation probability maps. Before being fed into the lesion segmentation model, the pathological RGB images are normalized by dividing the pixel values by 255, bringing them into the range of 0 to 1, which is consistent with the villous probability values. Then, this normalized RGB image and the villous probability map (as a fourth channel) are combined into a four-channel input image, which is then passed into the lesion segmentation model. This model is used to segment edema and proliferative lesions, and the corresponding process is shown in Fig. 2b.

In terms of network architecture, each segmentation model consists of three main components: the encoder model, the decoder model, and the segmentation head, as shown in Fig. 1b. The encoder model is responsible for extracting features from the input image and generating multi-dimensional feature maps. The decoder model then merges multi-scale features to form decoded feature maps. These decoded feature maps are passed to the downstream segmentation head for precise lesion area segmentation [31]. In this project, the villous segmentation network uses a separate encoder model, decoder, and segmentation head, while the edema and proliferative lesion segmentation tasks share the same encoder and decoder models, but have their own independent segmentation heads.

We have innovatively developed a method based on a multi-scale attention mechanism to effectively extract multi-scale features from pathology images. This helps us address two key issues: first, the significant size differences of lesions in the images; and second, the variations in the field of view and magnification of pathology slides and microscopes. These two issues make it difficult for traditional single-scale methods to handle all lesions and information effectively. To tackle these challenges, the pathology image is divided into patches of different scales, with the central region's patch set as the baseline, and patches of other scales are used to introduce multi-scale information. An attention mechanism is applied to each scale of patches to generate feature encodings, and during the feature fusion process, the query vector comes from a predefined intermediate-scale patch, while the key and value vectors come from patches of different scales. This design allows the model to integrate multi-scale features, capturing both macro and micro information from the pathology images, thereby better adapting to the features of lesions of varying sizes and effectively addressing the information differences caused by variations in field of view and magnification. Finally, we chose MaskFormerV2 as the segmentation head [32,33], which is highly consistent with the encoder-decoder architecture and performs excellently in lesion segmentation tasks, making it an ideal choice.

During training, the model adopts a two-stage training strategy. In the first stage, we freeze the parameters of the pre-trained feature extraction encoder and focus on training the downstream decoder model and segmentation head. To train this model, we used 40% of the dataset, which consists of 415 full pathological slides. In this stage, we tested different training strategies and compared them with other mainstream segmentation models, including U-Net++, FPN, and DeepLabV3+[18,34-36]. All models were trained on the same dataset, and we compared their average processing time and lesion segmentation accuracy on a single image. The results showed that the ViT-based segmentation head, MaskFormerV2, performed similarly in terms of processing time to the other models, with an average processing time of 85.6ms, compared to 95.7ms for U-Net++, 90.7ms for FPN, and 82.3ms for DeepLabV3+. In terms of accuracy, the ViT-based model achieved a pixel-level mIoU of 81.5%, whereas U-Net++ had an mIoU of 75.6%, FPN 74.5%, and DeepLabV3+ 75.9% (Supplementary Fig. 6, Supplementary Table 1). These results suggest that our ViT-based segmentation head demonstrates higher accuracy in lesion segmentation tasks, and since the feature extraction encoder also uses ViT, the overall network has an advantage in global training and fine-tuning processes.

We also tested the impact of different patch overlap rates on the model's performance, including 0%, 25%, 50%, and 75%. The experimental results showed that a 75% overlap rate achieved the best performance on the validation set, with mIoU improving from 81.5% at 0% overlap to 85.6% at 75% overlap (Supplementary Table 2). However, considering that excessive overlap may lead to overfitting, we ultimately chose a 50% overlap rate for training, as it provided a good balance between performance improvement and overfitting prevention. Additionally, due to the multi-scale nature of lesions in gestational trophoblastic diseases, we also tested the effect of single-scale training and multi-scale training with factors of 1.0, 1.2, and 0.8. The results showed that the multi-scale model achieved a pixel-level mIoU of 86.3%, outperforming the single-scale training model, which achieved 81.5%. This further validates that training with multi-scale pyramid images can improve the model's accuracy across different magnifications.

In the second stage of training, we unfroze the feature extraction encoder and used 60% of the training dataset, which consisted of 623 pathological slides, for global training, allowing the network to adjust all parameters. This stage significantly improved segmentation accuracy, especially in detecting small lesions, which resulted in better performance.

Initially, we used the pixel-level BCEWithLogitsLoss as our loss function. However, we still observed some issues. First, because the blank area contains more samples, while the edema, proliferative, and villous areas have relatively fewer samples, the model's performance on these areas was poor, especially in the proliferative regions, which occupy the smallest area. This is a class imbalance problem. To address this, we employed a weighted loss method, providing higher weights to the proliferative lesion areas:

$$L_{pixel} = -\frac{1}{N}\sum_{i=1}^{N}[w_1 \cdot y_i \cdot \log(\sigma(x_i)) + w_0 \cdot (1 - y_i) \cdot \log(1 - \sigma(x_i))] \quad (1)$$

Additionally, the BCEWithLogitsLoss primarily optimizes pixel-level accuracy but does not adequately consider lesion-level accuracy, particularly for detecting small lesions. The identification of small lesions is crucial for pathologists, yet they are often overlooked. Therefore, improving lesion-level accuracy is especially important for AI-assisted diagnostic systems.

To solve these two issues, we designed a hybrid loss function combining the pixel-level BCEWithLogitsLoss and lesion-level Dice coefficient loss. Specifically, the total loss function is defined as:

$$L_{total} = \lambda_1 \cdot L_{pixed} + \lambda_2 \cdot L_{lesion} \quad (2)$$

Where $L_{pixal}$ represents the BCEWithLogitsLoss, which computes the difference between the predicted and true values for each pixel, and $L_{lesion}$ is the lesion-level loss, using the Dice coefficient to measure the overlap between the predicted lesion region and the true lesion region[38,39]. The Dice coefficient loss is computed as:

$$L_{lesion} = 1 - \frac{|\widehat{Y}|+|Y|}{2\cdot|\widehat{Y}\cap Y|} \tag{3}$$

where $Y$ and $\widehat{Y}$ are the predicted and true lesion areas, and $|\cdot|$ denotes the area (number of pixels) of the lesion. This loss function helps the model improve segmentation accuracy for small lesions, which is particularly important for clinical diagnosis. By adjusting the weights $\lambda_1$ and $\lambda_2$, we were able to fine-tune the model to prioritize both pixel-level segmentation accuracy and lesion-level segmentation accuracy.

Through the implementation of this hybrid loss function, we were able to significantly improve the model's ability to detect small and proliferative lesions, addressing the challenges posed by class imbalance and small lesion detection. The final model demonstrated strong feature extraction and segmentation capabilities, achieving robust performance across different lesion types, making it an effective tool for assisting pathologists in diagnosing gestational trophoblastic diseases.

## GTDoctor - an Expert Model for GTD Diagnosis

The visual model provides doctors with image modality analysis results, including extracted multi-dimensional features and lesion segmentation results. Based on this foundation, we have designed a decision and analysis model that combines image analysis results with textual information, such as patient case details, to provide comprehensive diagnostic results and pathological analysis. Together, they form the GTDoctor's auxiliary diagnostic process. The structure of the GTDoctor model is illustrated in Fig. 1a. It provides a final case-level diagnosis based on the analysis of pathological slide images and the patient's basic information, including diagnostic opinions and detailed pathological analysis. GTDoctor can integrate diagnostic and analysis results and generate diagnostic reports.

For the decision and analysis model, on one hand, it needs to provide pathologists with a judgment on the patient's condition, resulting in a definitive disease classification. On the other hand, the pathological analysis should be as comprehensive and detailed as possible, offering personalized recommendations based on the case information, which can provide doctors with more information to explain the condition to patients and consider treatment strategies. Therefore, it is necessary to combine structured and unstructured results to provide doctors with more comprehensive diagnostic support. To this end, we propose a new decision analysis framework, including a random forest model based on hybrid features and a directionally trained large language model.

The random forest model analyzes and judges the features extracted by the visual model, combining the diagnostic experience of professional pathologists to output structured judgment results, including "Hydatidiform mole", "Choriocarcinoma" or "Normal abortion". In the feature selection phase, we combine manually selected artificial features with automatically extracted implicit features. The segmentation head output of the visual model provides segmentation results for different lesions. Based on these results, we collaborated with clinical doctors to select seven key artificial features: the number of edematous lesions, the number of hyperplastic lesions, the number of villi, the area ratio of edematous lesions, the area ratio of hyperplastic lesions, the proportion of abnormal villi, and the proportion of non-blank areas. These features directly reflect key information in the pathological slides, providing a basis for classification decisions. The encoder of the visual model outputs a 768-dimensional feature map, which contains implicit features beneficial for analysis. We use Principal Component Analysis (PCA) (Supplementary Fig. 7) to reduce the dimensionality of the high-dimensional feature map, finally forming a three-channel feature heatmap (as shown in Fig. 2d) [40]. Through methods such as heatmap region statistics and shape factor analysis, we extracted 15 potential features for classification. The total of 22 features we extracted combines the clinical expertise of pathologists with the potential patterns discovered by machine learning models through image analysis. Compared to relying on a single source of features, this hybrid feature selection method fully integrates interpretability and effectiveness, providing a more comprehensive basis for classification. During the training process, we used the training set from Cohort A-1. In terms of model structure, we used 20 CART decision trees with random feature selection, adopting the Bootstrap Aggregating (Bagging) integration strategy. Each decision tree's sample selection was random sampling with replacement, and the final classification result was determined by voting.

We utilized the large language model ChatGPT-4 as the foundational framework for case analysis, capitalizing on its robust capabilities in instruction interpretation and text comprehension. To augment its specialized proficiency in pathological diagnosis, we implemented a retrieval-augmented generation (RAG) approach to provide the model with supplementary theoretical knowledge[41]. We developed a high-quality, structured, and comprehensive literature database on gestational trophoblastic diseases, encompassing the latest research findings and authoritative guidelines pertaining to these conditions. This database comprises 45 key resources, including "Guidelines for the Diagnosis and Treatment of Gestational Trophoblastic Diseases" and "MRI Diagnosis and Pathological Correlation of Gestational Trophoblastic Diseases," which furnish the model with a comprehensive theoretical foundation and practical guidance. This ensures that the model integrates the most advanced medical knowledge when addressing diagnostic and therapeutic challenges. During the auxiliary diagnosis process, the model retrieves highly relevant information from the database based on the fundamental case details and lesion characteristics, selecting this as the training corpus for the large language model, thereby specifically enhancing its expertise in the field of pathology (Supplementary Fig. 5).

The decision analysis framework we designed can combine the characteristics of structured and unstructured data, make judgments according to the logic of doctors' diagnostic experience, and provide in-depth interpretation of diagnostic results based on the latest medical research. This report provides strong decision support for clinicians and helps patients better understand their condition (Supplementary Fig. 4).

## GTDiagnosis - an Intelligent Diagnostic Software System

To better serve hospitals and specialists in different settings, we have developed a comprehensive auxiliary diagnostic software system for gestational trophoblastic diseases based on GTDoctor. We named it GTDiagnosis, with GTDoctor as its core. GTDiagnosis supports both English and Chinese, and the system interface is shown in Supplementary Fig. 2 and Supplementary Fig. 3. The system has three main features.

1) It supports multi-scenario applications. On one hand, our model can analyze whole slide images (WSI) obtained from digital pathology slide scanners. On the other hand, for hospitals in areas with less advanced medical capabilities, we provide a microscope-based diagnostic solution. Our system can capture real-time microscope field views and perform lesion identification and segmentation, displaying the multi-lesion label information in real-time within the system. Furthermore, we developed a microscope historical field stitching module based on Fourier spectral transformation algorithms, which can intelligently stitch historical fields under the microscope and save the lesion segmentation results for further analysis, providing physicians with diagnostic assistance.

2) It has the ability to evolve through online learning during clinical use. We have implemented an online learning algorithm utilizing mini-batch stochastic gradient descent (SGD), paired with a model fusion module. This approach enables the model to update its parameters continuously with new data collected during clinical usage. The new model parameters are then temporally fused with the existing model in a weighted manner, ensuring that the model retains its robustness while incorporating the learning from the new data. Temporal fusion helps maintain the balance between preserving prior knowledge and adapting to new patterns, thereby enhancing both generalization performance and the stability of the analysis outcomes. This design allows for the integration of new and valuable samples encountered during diagnoses. Doctors can revise the AI-generated labels based on clinical feedback and use these updated samples as training data for the model. Through this positive feedback loop, the model continuously refines itself over time, improving accuracy and adaptability. This method not only ensures that the model remains robust to variations in clinical data but also accelerates the system's long-term development and performance.

3) It is user-friendly. We trained three pathologists on how to use GTDiagnosis, and they were able to master the system and use it proficiently in less than 20 minutes on average. We also conducted experiments to assess the impact of GTDiagnosis on the efficiency and accuracy of diagnoses, the results of which will be presented in a later section of this paper, and are very promising. Through the design of GTDiagnosis, we have provided a comprehensive, multi-dimensional solution for the auxiliary diagnosis of gestational trophoblastic diseases, improving the entire professional workflow from patient data import, large-model lesion segmentation and recognition, to intelligent decision-making, diagnosis, and pathology report generation. The usage process and application scenario of GTDiagnosis are shown in Supplementary Fig. 1.

## Acknowledgements

This work was supported by the Youth Project of Beijing Natural Science Foundation (Grant No. 4254099).

## Author contributions

Yuhang Liu, Li Shi, Wenge Que designed the research plan. Yueyang Cang, Xingtong Wang, Xiaoteng Zhang and Pingge Hu executed the research plan and developed the model algorithms. Xinru Bai, Kuisheng Chen, Peiyu Xu, Lixia Zhang, Xinmin Li, Jingya Li and Qiaoting Xie annotated the lesion regions and organized the data. Yuhang Liu and Yueyang Cang and Xingtong Wang conducted the experiments and wrote the manuscript. Li Shi, Wenge Que and Xianxu Zeng revised and edited the manuscript. All authors read and approved the final version of the article.

## References

References should be numbered sequentially first throughout the text, then in tables, followed by figures and, finally, boxes. References may only contain citations and should list only one publication with each number. Examples:

## Competing interests

The authors declare that they have no conflict of interest.

## Supplementary Information

Supplementary Information can be found in the compressed folder.

# Figures

**Fig. 1 Diagram of the overall structure of GTDoctor**

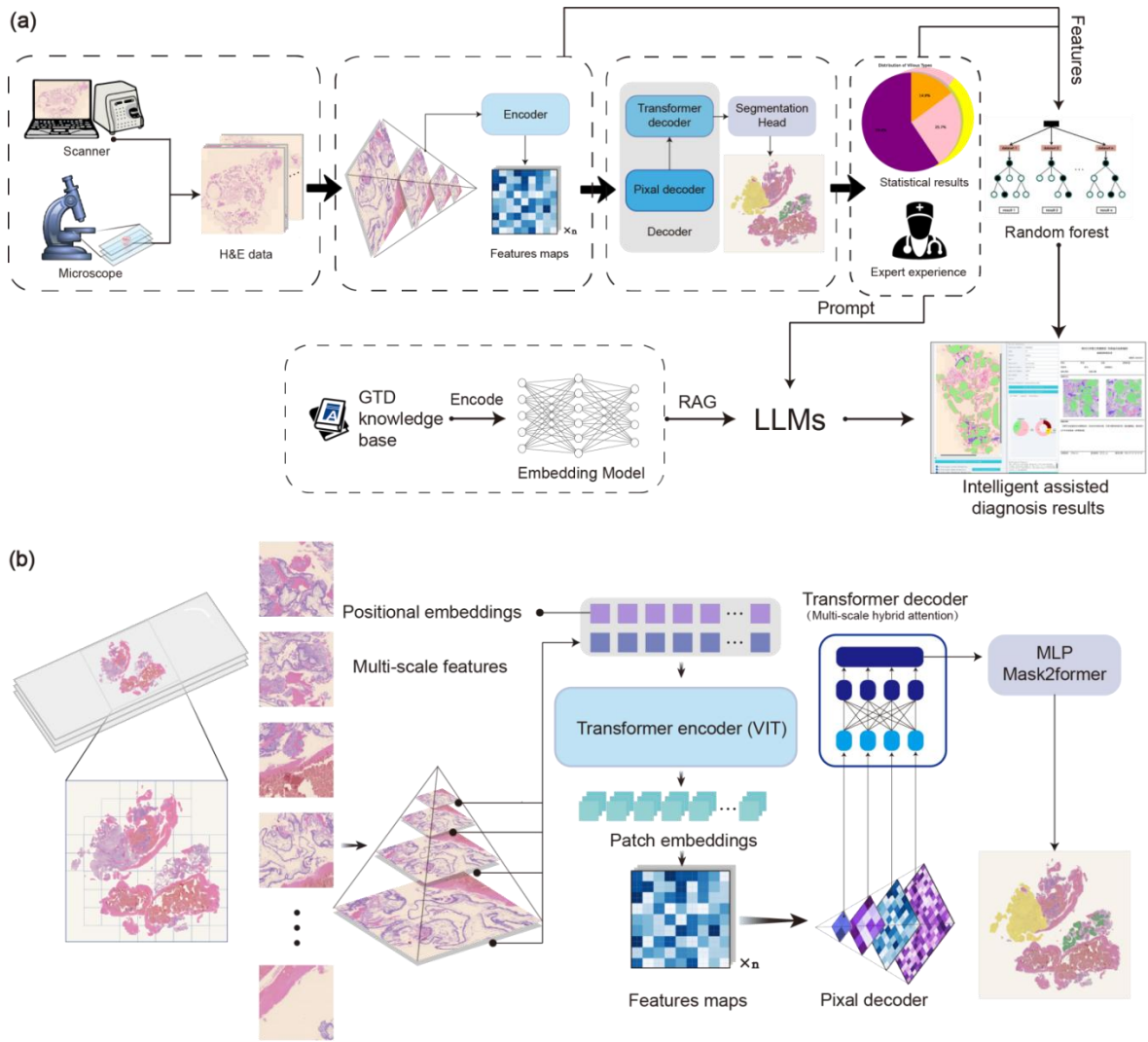

**a** The workflow of GTDoctor. The acquired H&E data is processed through the visual model for encoding, decoding, and segmentation. The segmentation results undergo statistical analysis and manual feature extraction. The output from the encoder also serves as part of the feature set input into a random forest decision model for diagnosis. Additionally, patient information and pathology slide details are provided to a domain-specific large language model for diagnostic analysis. Together, these components contribute to the final analysis results. **b** Algorithm structure of the Lesion Segmentation Vision Model, which captures lesion information at different scales through multi-scale feature encoding and hybrid attention mechanisms. AI: Artificial Intelligence. GTD: Gestational Trophoblastic Diseas. H&E: hematoxylin-eosin staining.

**Fig. 2 Sample and workflow of the visual model**

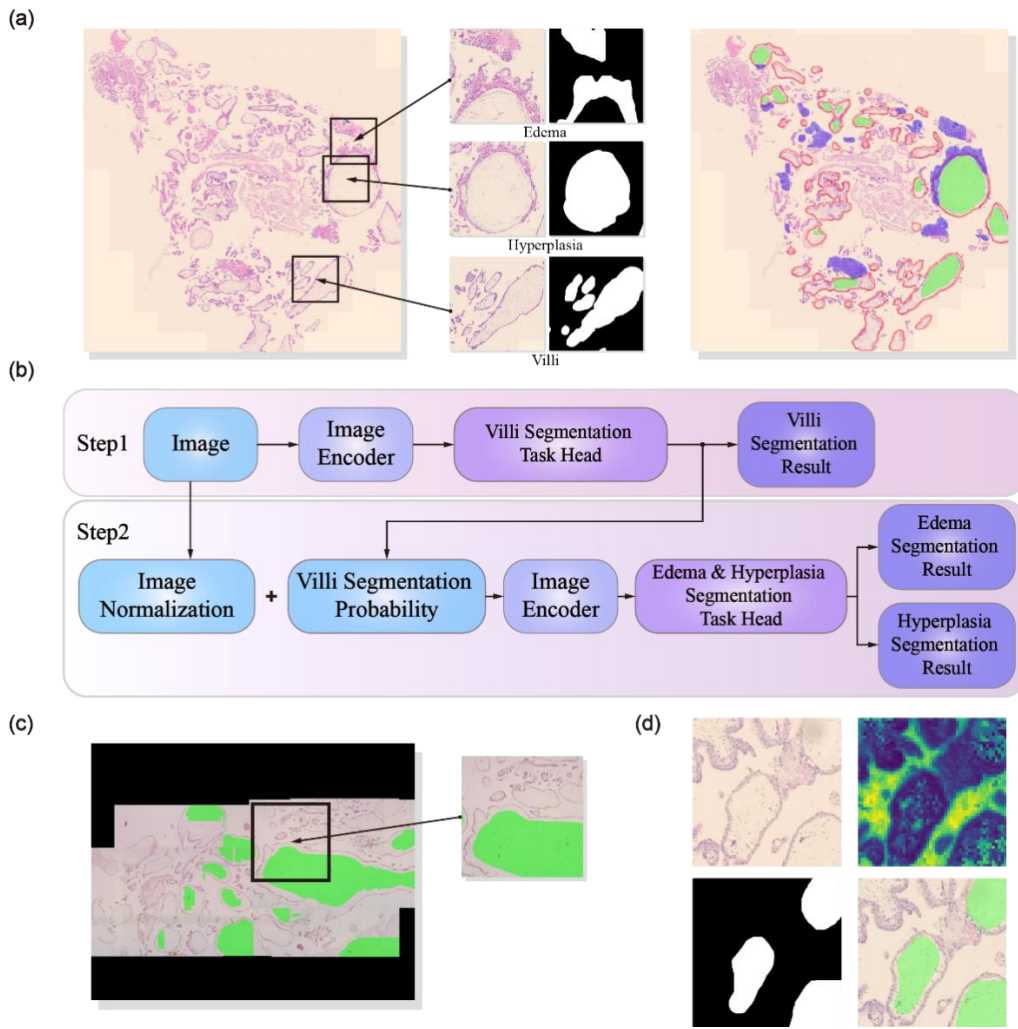

**a** Edema, Hyperplasia, and Villi regions in the Whole Slide Image, with the annotated image shown on the right. **b** The lesion segmentation process based on two stages. **c** The field of view of the auxiliary diagnostic system under the microscope and the result after field stitching. **d** From top left to bottom right: the original pathology slide, the dimension-reduced feature heatmap, the segmented mask, and the final lesion display image from the auxiliary system.

**Fig. 3 Performance of GTDiagnosis in different tests.**

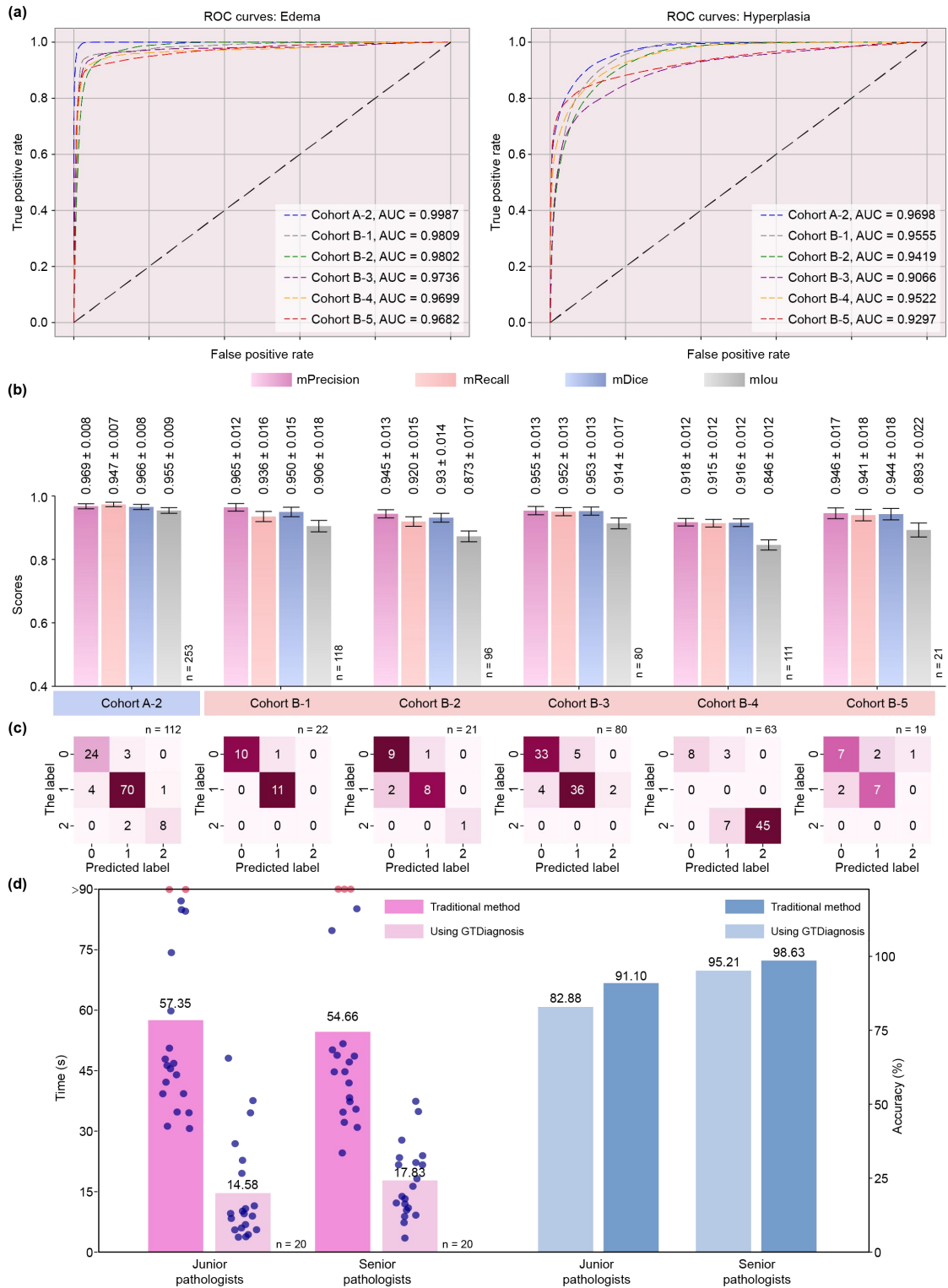

**a** ROC curves for pixel-level edema and hyperplasia lesion segmentation results (cohort A-2: internal test set, cohort B-1 to B5: external test sets). **b** The mprecision, mrecall, mdice, and miou scores of segmentation results (cohort A-2: internal test set, cohort B-1 to B5: external test sets). **c** Confusion matrix for disease diagnosis results across different cohorts, where 0 represents Normal abortion, 1 represents Hydatidiform mole, and 2 represents Choriocarcinoma (cohort A-2: internal test set, cohort B-1 to B5: external test sets). **d** The effect of GTDiagnosis on the average diagnostic time and diagnostic accuracy for junior and senior pathologists (cohort C-1).

Tables

**Table 1 Baseline characteristics of patients included in this study (n=831 cases)**

| Characteristic | | Total number |
|---|---|---|
| **Age (years)** | | |
| | ⩽19 | 12 (1.4%) |
| | 20-29 | 281 (33.8%) |
| | 30-39 | 435 (52.4%) |
| | 40-49 | 87 (10.5%) |
| | 50-59 | 16 (1.9%) |
| **Gestational Weeks (weeks)** | | |
| | ⩽4 | 11 (1.3%) |
| | 5-7 | 339 (40.8%) |
| | 8-10 | 223 (26.8%) |
| | 11-13 | 46 (5.5%) |
| | 14-16 | 7 (0.9%) |
| | Unknow | 205 (24.7%) |
| **Categories** | | |
| | Normal abortion | 165 (19.9%) |
| | Hydatidiform mole | 602 (72.4%) |
| | Choriocarcinoma | 64 (7.7%) |